\documentclass[twoside,11pt]{article}
\usepackage{jair}
\usepackage{graphicx}
\usepackage{subcaption}
\usepackage{arydshln}
\usepackage{amsmath}
\DeclareMathOperator*{\argmax}{arg\,max}
\DeclareMathOperator*{\argmin}{arg\,min}
\usepackage{multirow}
\usepackage{xcolor}
\usepackage{booktabs}

\usepackage[backend=biber,style=apa,natbib=true]{biblatex}
\usepackage[american]{babel}
\DeclareLanguageMapping{american}{american-apa}
\DeclareFieldFormat{apacase}{#1}

\usepackage{hyperref}
\hypersetup{
    colorlinks=true,
    linkcolor=blue,
    filecolor=blue,      
    urlcolor=blue,
    citecolor=blue
}
 
\ShortHeadings{Transportation Behavior Influence}
{Mohan, Rakha, \& Klenk}
\firstpageno{1}

\begin{document}

\title{Acceptable Planning: Influencing Individual Behavior to Reduce Transportation Energy Expenditure of a City}

\author{\name Shiwali Mohan \email shiwali.mohan@parc.com \\
       \addr Palo Alto Research Center, Mail Stop: 3333\\
       Coyote Hill Road, Palo Alto, CA  94034 USA \\ \\
       \name Hesham Rakha \email hrakha@vtti.vt.edu\\
        \addr Virginia Tech Transportation Institute, Mail Stop: 3500\\
        Transportation Research Plaza Blacksburg, VA 24061 \\ \\
\name Matthew Klenk \email matthew.klenk@parc.com \\
       \addr Palo Alto Research Center, Mail Stop: 3333\\
       Coyote Hill Road, Palo Alto, CA  94034 USA}


\maketitle

\begin{abstract}
Our research aims at developing intelligent systems to reduce the transportation-related energy expenditure of a large city by influencing individual behavior. We introduce \textsc{Copter} - an intelligent travel assistant that evaluates multi-modal travel alternatives to find a plan that is \emph{acceptable} to a person given their context and preferences. We propose a formulation for \emph{acceptable} planning that brings together ideas from AI, machine learning, and economics. This formulation has been incorporated in \textsc{Copter} that produces acceptable plans in real-time. We adopt a novel empirical evaluation framework that combines human decision data with a high fidelity multi-modal transportation simulation to demonstrate a $4\%$ energy reduction and $20\%$ delay reduction in a realistic deployment scenario in Los Angeles, California, USA.
\end{abstract}

\section{Introduction}
\label{Introduction}

Transportation is one of the largest consumers of energy in the world - in the United States, it accounted for $29\%$ of energy consumption in $2016$ \citep{eia2016}. Despite occupying a central place in the economy, transportation is far from efficient. Many areas of urban transportation networks are underutilized while other areas are congested. Congestion alone in the United States wastes $6.9$ billion hours and $3.1$ billion gallons of fuel per year \citep{schrank2015}. Thus, transportation has become an important policy \& innovation problem. Recently, public and private entities have begun introducing new transportation services including bike/scooter share, car share, ride hailing, and carpooling to complement private vehicles and existing public transit. This increasing accessibility of novel modes and the corresponding need for planning solutions creates an opportunity for \emph{mobility marketplaces} that aggregate a variety of offerings into a single market, similar to what Travelocity or Expedia does for air travel. Such mobility marketplaces enable the creation of new technologies to assist travelers in making decisions that meet their needs and benefit their community.

Our long-term research goal is to develop AI methods to support efficient transportation decision-making reducing overall energy consumption and increasing citizen satisfaction in metropolitan areas. There are two important components to changing the transportation behavior of a large population toward this \emph{social good} goal. The first - the \emph{multi-modal route planning} problem - addresses how to reason about the large decision space of a variety of modes, routes, and combinations that are accessible to a person to determine an energy-efficient trip. The second component - the \emph{influence} problem - is guiding an individual's behavior such that they actually adopt trips that reduce the overall energy expenditure of a region. While the multi-modal route planning problem has been studied in detail by the AI community \citep{bast2016route}, the influence problem is largely unexplored in the AI literature. When influence is considered, it is from the perspective of monetary incentives \citep{doi:10.1177/0361198118775875,azevedo2018tripod}, which poses a business challenge. Here we focus on understanding the factors that underlie transportation decision-making to enable influence without monetary incentives.

Our research on AI systems that act to influence human behavior towards a social good goal is part of a larger research agenda on design and analysis of \emph{human-aware} AI systems. Often while designing AI systems, little attention is devoted to modeling the behavior of humans who invariably are crucial decision makers. Human-aware AI systems \citep{khampapati2018} seek to address this gap and pose modeling humans as a question central to AI system design. While this is a worthwhile pursuit, current human modeling methods in AI are extremely limited in their scope \citep{albrecht2018autonomous}. Largely, human modeling research in AI originates in game domains where humans are modeled either as collaborators or opponents with fixed objective functions. These models are not sufficient for representing the causation of dynamic and evolving human behavior in the real-world (such as choosing a mode of transport) and consequently, cannot be applied to influence it.

Our approach to building human-aware AI systems is to borrow models and methods from social sciences, in particular, behavioral economics, and explore how they can be integrated in computational systems in a principled fashion. Additionally, we relinquish typical AI metrics of accuracy and efficiency and define new \emph{human-centered} desiderata for AI systems to ensure that we are making progress towards systems that are truly human-aware. In our problem, the desiderata is related to \textit{influence} - how likely is a person to change their behavior when provided with a recommendation.
To develop a computational formulation of \textit{influence}, we engage with research in economics on rational choice and human preference modeling. Further, we adopt empirical methods prevalent in social science research to evaluate our proposed formulation. Then, we engage with transportation modeling research to simulate the impact of applying our approach on the energy consumption of a large city. By engaging these disciplines, we propose a comprehensive methodology for developing intelligent solutions for social good. More specifically, our paper makes the following contributions:
\begin{enumerate}
    \item Introduces the transportation influence problem for social good and a useful component - \textit{acceptability},
    \item Develops a machine learning approach to estimate acceptability,
    \item Evaluates the formulation of acceptability using a human participant study,
    \item Proposes and implements a novel integration of AI planning theory and economic choice theory for \emph{acceptable planning}, and
    \item Evaluates the integrated approach by combining human decision data with a high fidelity transportation simulation model of the Los Angeles area.
\end{enumerate}

\section{The Traveller Influence Problem}
\label{sec:influence_problem}
Social sciences research proposes several factors that are useful for influencing someone to change their behavior. The transportation mode choice framework \citep{ben1985discrete} suggests that attributes that define the context in which the trip is undertaken, such as distance of the trip, cost of taking the mode, duration of travel etc. play a role in people's preference for a specific mode. Recommending a mode that is higher on a person's preference list may increase the likelihood that the recommendation will be adopted. Just-in-time adaptive interventions for health behavior change \citep{nahum2015building} suggest that recommendations provided at opportune times are more likely to influence behavior. Further, research in persuasive messaging \citep{benoit2008persuasive} has studied the impact of various types of message framing - how information is conveyed - has on individuals in influencing their choices and behaviors. Building on these results, we define the travel influence problem as delivering an \emph{acceptable}, \emph{timely}, and \emph{compelling} travel recommendation that results in a person selecting a sustainable mode.

\begin{figure}
  \centering
    \includegraphics[width=1\textwidth]{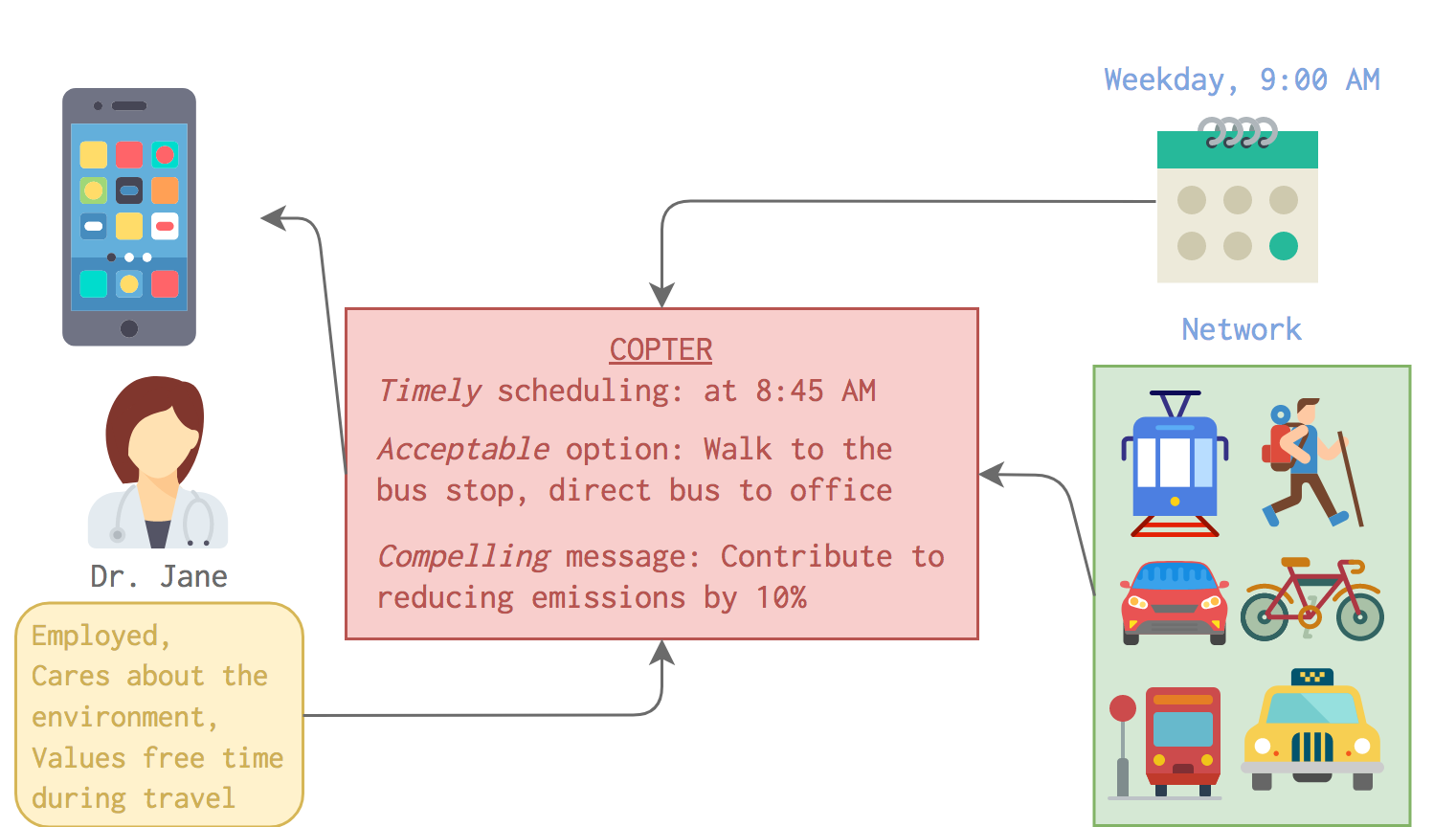}
    \caption{The transportation influence problem seeks to alter traveler behavior by suggesting routes based on the available options, the dynamic context including trip purpose, and a model of the traveler's preferences.}
    \label{fig:arch}
\end{figure}

Our approach to the influence problem is embedded within an intelligent \emph{travel assistant} - Collaborative Optimization and Planning for Transportation Energy Reduction (\textsc{Copter}) and aims at operationalizing influence strategies for travellers. Figure \ref{fig:arch} contains a functional description of the system and we illustrate \textsc{Copter's} desired influence behavior with the following example. Assume that Jane makes a regular trip driving to her office every weekday morning and has \textsc{Copter} installed on her smartphone. In a \emph{timely} fashion, $15$ minutes before she has to leave, \textsc{Copter} suggests an alternative trip - walk to the bus stop and take the direct bus to her office - chosen from all alternatives available to Jane. This trip is not only energy-efficient, but also \textit{acceptable} to Jane given the context of her travel and her preferences. \textsc{Copter} makes this suggestion by considering that the route is direct, that a lot of people in her neighborhood take the bus, and that she is traveling to her workplace and therefore, will not be carrying heavy loads. Further, \textsc{Copter} recognizes that she cares about her impact on the environment and frames this information in a \emph{compelling} way by telling her the emissions she can reduce by taking this trip. 

This paper explores how acceptability of a mode can be defined for transportation planning and how acceptability affects the adoption of a recommendation. The remainder of the paper is organized as follows. We begin by describing the multi-modal planning framework and choice theory framework that form the basis of our approach, in Section \ref{sec:preliminaries}. Then, in Section \ref{sec:defining-acceptability}, we formally define the notion of \textit{acceptability} for planning. We develop a machine learning method to estimate individual acceptability from a population-level, transportation behavior dataset from Los Angeles, California in Section \ref{sec:ml_approach}. We show that this acceptability model is predictive of adoption through a human participant study in Section \ref{sec:evaluation_acceptability}. Upon validating our definition of acceptability, we incorporate it in the planning framework in Section \ref{sec:acceptable_planning}. Finally, we demonstrate the energy impact of our approach using a high-fidelity simulation model of transportation energy integrated with human behavior in Section \ref{sec:experiments}. In Section \ref{sec:related_work}, we study research threads pursued by other researchers from AI, transportation, and behavior change. We conclude with a discussion of the future work in Section \ref{sec:conclusion}.

\section{Preliminaries}
\label{sec:preliminaries}
To understand the novelty of our approach, we first review the two strands of research we build upon, and later, we describe their limitations for the transportation influence problem. 

\subsection{Multi-Modal Route Planning}
Routing in transportation networks is a well studied problem in computer science dating back to Dijkstra's algorithm \citep{dijkstra1959note}. Today, through pre-computation and heuristics, planners identify optimal routes for continent-size networks while guaranteeing optimal solutions in fractions of seconds \citep{bast2016route}. 

These approaches use the standard transportation network representation as a directed graph $G = (V,E)$, where $V$ is a set of nodes and $E \subseteq V \times V$ is a set of directed edges. The formalism also includes a standard representation of time, $T$. A request for transporting a person from their origin location $v \in V$ starting at a certain time $t_s \in T$ to their destination $w \in V$ before a certain deadline $t_e \in T$ is formalized as $q = (v,w,t_s,t_e)$. The planning problem is formalized as $(G, q)$. Traversal of each edge is represented as a pair $(e,t)$ where a person begins moving along the edge $e$ at time $t$. A plan $\pi = ((e_1,t_1),...,(e_n,t_n))$ is a valid solution of the problem $(G, q)$ if and only if: 
\begin{enumerate}
\item the edges connect the origin with the destination. $e_1 = (v_0, v_1),e_2 = (v_1,v_2),...,e_n = (v_{n-1},v_n) \wedge (v_0 = v) \wedge (v_n = w)$, 
\item temporal constraints are satisfied, $t_s \leq t_1 + dur(e_1) \leq t_2 + dur(e_2) \leq ... \leq t_n + dur(e_n) \leq t_e$ where $dur(e)$ represents the time it takes to traverse the edge $e$. 
\end{enumerate}

To find an optimal plan, cost function $cost(e)$ is defined for every edge in $G$. An optimal plan is a plan that minimizes the cumulative cost 
$\pi^* = \argmin_{\pi \in \Pi} \sum_{e \in \pi} cost(e_i)$, where $\Pi$ is the set of all valid plans.
In the simple case, the cost is the time it takes to traverse the edge, $cost(e_i) = dur(e_i)$. $A^*$ search with a reasonable heuristic (e.g., euclidean distance) efficiently solves these problems.

Recent work provides extensions for multi-modal trips \citep{Dibbelt:2015:UMR:2627368.2699886} and more elaborate cost functions.  This is achieved by adding additional edges between vertices corresponding to the various modes (e.g.,  public transportation, biking, walking) that are available to traverse them. When the edges represent public transit vehicles, they have temporal constraints representing the vehicle schedules. Each edge $e$ is labeled with the mode it represents, e.g., $lbl(e) = b$ for an edge representing taking the bus. Consequently, each plan $\pi$ corresponds to a sequence of labels, \emph{word} ($w_p$). This imposes a third  requirement on valid solutions that the word is part of the language of acceptable mode sequences ($w_p \in L$), represented using regular expressions. This constraint eliminates plans that are impossible, such as taking a bus and then getting on a bike. Others \citep{delling2013computing,botea2016hedging,dvorak2018} have shown that in this formulation, the cost function can consider additional dimensions including price, energy, and schedule risk. The cost function is formalized as $cost(\pi) = \sum_{e \in \pi}\sum_{\varphi \in \Phi}\theta_{\varphi} \varphi(e)$ where $\Phi$ is the set of evaluative functions including energy consumed, price, etc. and $\theta_{\varphi}$ corresponds to how much importance that evaluative function has toward computing the overall edge cost. 

To support personalization, numerous researchers have extended the basic notion of an edge-based objective function to capture more information.  For example, deployed planners now combine travel time, cost, and convenience measures in selecting multi-modal plans \citep{delling2013computing}.  In the research setting, methods for incorporating even more criteria (e.g., robustness to missing connections, \cite{botea2016hedging}, safety, \cite{Liu2018}) have been used. 

\subsection{Utility-Based, Probabilistic Choice Theory}
Economists use the framework of rational choice theory \citep{tversky1986rational} to study how humans make decisions from a set of competing alternatives. Empirical research in transportation economics \citep{domencich1975urban} has shown that such frameworks are useful in predicting transportation mode shares in urban populations. The theory posits a choice process: for a trip, the individual first determines the available mode alternatives; next, evaluates the attributes of each alternative relevant to the travel decision, and then, uses a decision rule to select an alternative. The rational decision-making perspective motivates \emph{utility maximization} as a relevant decision rule. Utility maximization is based on two fundamental concepts. First, the attributes of a mode alternative can be reduced to a scalar utility value. This assumption implies a compensatory decision process; i.e, the decision maker will make \emph{trade-offs} among the various attributes. For example, an individual may choose a pricier mode if the comfort provided by that mode compensates for the increased cost. Second, the decision maker selects the alternative with the highest utility value. This framework can be expressed as:
\[
util(x_i, f_p) \geq util(x_j, f_p) \forall j \implies i \succ j \forall j \in C 
\]
where $util()$ is the utility function, $x_i, x_j$ are vectors of attributes describing mode alternatives $i$ and $j$ (e.g, travel time, travel cost, etc.), $f_p$ is a vector of characteristics of the decision maker that influence their preferences (e.g., income, number of automobiles), $i \succ j$ means that $i$ is preferred to $j$, and $C$ is the choice set of various alternatives. 

The deterministic theory assumes all aspects of the decision process can be observed and measured. Transportation decision-making is only partially-observable. By adding noise, the utility of an alternative for an individual can be split into two components - observable utility and error:
\[
util(x_i, f_p) = val(x_i, f_p) + \epsilon(x_i, f_p)
\]
Here $val(x_i, f_p)$ is the observable portion of the utility. This quantity is modeled as a linear combination of measurable attributes such as travel time, travel cost, walk distance, etc: 
\begin{equation}
\label{eq:value}
\begin{split}
val(x_i, p) = \gamma_1 \times x_{i1} + \gamma_2 \times x_{i2} + ... + \gamma_k \times x_{ik} + \\ \lambda_1\times f_{p1} + \lambda_2 \times f_{p2} + ... + \lambda_l \times f_{pl}
\end{split}
\end{equation}

$\gamma_k$ is the parameter which defines the direction and importance of the attribute $k$ on the utility of an alternative $i$. Similarly, $\lambda_l$ defines the direction and importance of the person's characteristic $l$ .

$\epsilon(x_i, f_p)$ is the unobservable portion - the error term - of the utility function. Prior work \citep{koppelman2006self} in transportation mode choice recommends an assumption that the error terms of various alternatives are independent. This assumption leads to a \emph{multinomial logit model}. This model estimates the probability $Pr(i, p)$ of the person $p$ selecting the alternative $i$ from the set of alternatives $C$:
\begin{equation}
\label{eq:probability}
Pr(i, p) = \frac{e^{val(x_i, f_p)}}{\sum_{j \in C} e^{val(x_j, f_p)}}
\end{equation}
Empirical work in transportation research \citep{ben1985discrete} has studied how $\gamma$ and $\lambda$ parameters in equation \ref{eq:value} can be estimated from human choice data using maximum likelihood estimation given equation \ref{eq:probability}. 

\subsection{Limitations \& Challenges for Influence}
While each of these threads provides insights for the influence problem, there are significant gaps.
While we incorporate many of the ideas from AI planning including constraints on multi-modal transit graphs and time-dependent edges, finding optimal routes depends on computing costs for traversing each edge in the transportation network which poses two challenges in our context: (1) They include a large number of user parameters that are difficult to elicit from users (e.g., how much do you value time riding a bus versus a train versus driving a car?), and (2) Empirical work from transportation choice theory indicates that there are attributes that do not correspond to edge weights (e.g., household income, number of vehicles at home). We are unable to apply choice theory directly, because by itself this theory does not suggest how mode alternatives can be generated.

Furthermore, neither of these methods explicitly provide a definition of \emph{acceptability}, \emph{adoption}, or influence. Planning computes the optimal path for an individual given a cost function defined for every edge and choice theory predicts the probability of a trip occurring given some observable characteristics about the person and the mode of transport. 
The results of these computations are not the same as an estimate of the likelihood that the user will take an alternative mode if recommended (mode adoption likelihood). This is necessary for a solution to our decision-theoretic problem.
That is, in \textsc{Copter}, each traveler has an expected plan for an upcoming trip, and our goal is to produce an \emph{acceptable} alternative that changes their behavior to reduce energy consumption across the network. 
In the sections below, we propose how these challenges can be addressed for \emph{acceptable} planning.

\section{Defining Acceptability}
\label{sec:defining-acceptability}
We adapt choice theory's decision process to incorporate \emph{acceptability} as follows. Jane (from Section \ref{sec:influence_problem}) drives to her office every weekday on a certain route. The choice theory suggests that this usual route ($u$) has a certain measurable utility, $val(x_u, f_p)$ and a certain probability of being selected based on equation \ref{eq:probability}. Here $x_u$ is the vector of attributes representing Jane's usual mode such as time, distance, etc. and $f_p$ is a vector that describes Jane's attributes that pertain to mode selection such as income, education level etc. Upon receiving the recommendation, $r$, from \textsc{Copter}, Jane evaluates it against her usual means of travel. The recommendation $r$ has a measurable certain utility, $val(x_r, f_p)$,  given the utility function underlying Jane's preferences. On adopting the recommendation, Jane will experience a \emph{switching cost} - $\Delta_{u,r,p}$ - the difference in utilities of her usual route and the proposed route. Conversely, $\Delta_{r,u,p} = -\Delta_{u,r,p}$, can be understood as the \emph{switching gain} a person makes on adopting the recommendation $r$.

Following probabilistic choice theory (equation \ref{eq:probability}):
\begin{equation}
\label{eq:acceptability}
\begin{split}
\frac{e^{val(x_u, f_p)}}{e^{val(x_r, f_p)}}= \frac{Pr(u, p)}{Pr(r, p)}  \\
e^{val(x_u, f_p) - val(x_r, f_p)} = \frac{Pr(u, p)}{Pr(r, p)} \\
\Delta_{u,r,p} = val(x_u, f_p) - val(x_r, f_p)  = \ln{\frac{Pr(u, p)}{Pr(r, p)}} \\
\Delta_{r,u,p} = -\Delta_{u,r,p} = \ln{\frac{Pr(r, p)}{Pr(u, p)}} \\
\end{split}
\end{equation}
 
We postulate that the acceptability of a route is related to its switching gain. If we can estimate the likelihood of a person taking their usual mode of travel and that of the recommended mode, we can compute the change in utility and consequently estimate the recommendation's acceptability. This formulation is a significant departure from prior empirical work in transportation research. The prior work focuses on estimating the coefficients in equation \ref{eq:value} to compute exact utility values given measurements of relevant, observable attributes. Exact values of coefficients are useful for comparing various attributes in the model to understand the trade-offs a person may be willing to make and for computing monetary incentives. For influence, this exact computation of utility value is not necessary and the switching gain (and acceptability) can be computed from estimated likelihoods alone.

However, before acceptability is integrated in the planning framework, the following questions merit further exploration. 
\begin{itemize}
    \item What is the exact quantity that represents acceptability? Based on the discussion here, we begin by considering three candidate definitions:
        \begin{enumerate}
            \item switching gain: $\Delta_{r,u,p}$
            \item odds: $e^{\Delta_{r,u,p}} = Pr(r,p)/Pr(u,p)$
            \item likelihood of the recommended mode: $Pr(r,p)$ 
        \end{enumerate} 
        A related question is - how can acceptability be computed for a person when there is no prior observed mode selection data for them?
    \item Does acceptability of a mode affect its adoption when it is included in a recommendation? Are there other factors that may influence mode adoption above and beyond a recommendation's acceptability?
    
\end{itemize}
We answer these questions through an empirical approach described in the following sections. In Section \ref{sec:ml_approach}, we develop a machine learning approach using a large-scale, publicly available survey dataset to estimate mode likelihood for a person taking a given trip. These likelihoods can be used to compute acceptability using any candidate definition. Then, we compare different candidate definitions of acceptability (computed using these likelihoods) in Section \ref{sec:evaluation_acceptability} determine which definition best predicts adoption. 

\section{Estimating Acceptability: A Machine Learning Approach}
\label{sec:ml_approach}
As explained earlier, in order to estimate acceptability of a recommendation $r$ for a person $p$, we need to calculate the likelihood of taking the mode of the recommendation, $Pr(r, p)$ as well as for their usual mode, $Pr(u, p)$. Here we explore how a data-driven, machine learning method can be used to estimate these likelihoods. A typical ML classifier, such as a random forest classifier, generates a prediction of a class, given a set of input features, by computing probability for each class label and selecting the class with the highest probability. Thus, the classifier not only predicts the class label, but also the probability for every other class label. This probability estimate is critical for our approach. 

\subsection{Dataset}
To develop our ML classifiers, we used the California Household Travel Survey (CHTS; \cite{CHTS}) that consists of a single-day (distributed in a year) travel diary of people from $58$ counties of California and $3$ counties of Nevada. To ground our research within the context of a specific metropolitan area, we extracted trip data for $4,889$ Los Angeles (LA) residents belonging to $2,006$ households. Each sample in this dataset is a trip (defined by an origin-destination pair) taken by a specific individual. This dataset contains $78,380$ trips undertaken in Los Angeles county. For each trip, we extracted the following features:
\begin{itemize}
\item \textit{trip-related}: trip distance,
\item \textit{demographics}: such as education level, number of people/students/workers in the household,
\item \textit{employment}: number of hours worked every week, income, number of jobs, flexibility of work, 
\item \textit{mode accessibility}: number of automobiles in the household, number of bicycles in the household, owns a driver's license \& transit pass, and
\item \textit{mode experience}: transit used, bike trips made \& walking trips made in the past week.
\end{itemize}
These trips were made using $7$ different modes - walk, cycle, bus, subway/train, drive, auto passenger, and motorcycle.

\subsection{Models}
We explored two different multi-class prediction problems: mode prediction and category prediction. The mode prediction problem is predicting the most likely mode given some characteristics about the trip and the person making that trip. For category prediction, we created additional labels with walking \& cycle as \textit{non-motorized}, driving, riding, \& motorcycle as \textit{motorized}, and bus \& subway as \textit{public transport}. Consequently, the category prediction problem is predicting the category of the mode given trip and person characteristics. For each prediction problem, we developed two kinds of classifiers using standard open-source implementations:
\begin{enumerate}
    \item Random forest \citep{scikit-learn}: an ensemble of decision trees with $20$ estimators of $30$ depth each.
    \item Multi-layer perceptron  \citep{tensorflow2015-whitepaper}: a multi-layer perceptron with $4$ layers ($[1000, 500, 100, 100]$).
\end{enumerate}
We selected these classifiers because they impose minimal constraints on the relationships between features and class labels and can capture non-linear decision boundaries. As a starting point of exploring the prediction problems, these types of classifiers are fairly reliable. The specific parameters were found by a brief, bucketed linear hyperparameter search.

To study if these classifiers captured useful predictive information, we compared their performance with two baselines:
\begin{enumerate}
    \item \textit{most frequent}: This baseline assigns the most frequent class label to every sample.
    \item \textit{weighted random}: This baseline randomly assigns class labels to each sample based on its distributional frequency in the dataset.
\end{enumerate}

\subsection{Results}
\begin{table}[t]
\begin{center}
\def\arraystretch{1.1}
\begin{tabular}{lcccc}
\toprule
Mode & Baseline 1 & Baseline 2 & Random& Multi-layer \\
& (most frequent) & (weighted random) & Forest&Perceptron \\
\midrule
\textit{Walk} &0.00&0.12&0.82*&0.62 \\
\textit{Cycle} &0.00&0.00&0.81*&0.28 \\
\textit{Bus} &0.00&0.02&0.78*&0.38 \\
\textit{Subway/train} &0.00&0.00&0.58*&0.05 \\
\textit{Drive} &0.72&0.56&0.93*&0.86 \\
\textit{Ride }&0.00&0.28&0.84*&0.65 \\
\textit{Motorcycle} &0.00&0.00&0.80*&0.00 \\
\midrule[0.15pt]
\textbf{\emph{Total}} & 0.68& 0.40& 0.88*& 0.74\\
\bottomrule
\bottomrule
Mode category &&&&\\

\midrule
\emph{Non-motorized} &0.00&0.05&0.83*&0.60 \\
\emph{Public transit} &0.00&0.14&0.79*&0.43 \\
\emph{Motorized} &0.90&0.82&0.97*&0.93 \\
\midrule[0.15pt]
\textbf{\emph{Total}} &0.68&0.70&0.94*&0.86\\
\bottomrule
\end{tabular}
\end{center}
\caption{F1-scores for predicting mode classes and major modes with various machine learning methods. * indicate the best performing classifier for that mode.}
\label{table:prediction}
\end{table}

$F1$ scores for both classifiers as well as baselines on a $20\%$ test set are shown in Table \ref{table:prediction}. We see that both ML classifiers perform significantly better than the baselines for not only predicting each class (\& category) but have a better overall performance as well. These results suggests that the ML classifiers contain useful predictive information that can be used to estimate the switching gain (as well as acceptability of a recommendation). We can see that the random forest classifier has better performance than the multi-layer perceptron.\footnote{With additional effort and design, there is room for significant improvement in each of these classifiers. The results presented were sufficient for our objective of evaluating the overall framework.} This is expected due to the limited size of the dataset. We further see that the performance on category prediction is better than mode prediction. Again, this is to be expected as combining mode classes leads to more data to learn from and consequently better performance. Given that the random forest classifier achieves the best performance on this dataset, this classifier was selected for implementing the remainder of our approach.

\begin{figure}
    \begin{subfigure}{.5\linewidth}
    \includegraphics[width=1\textwidth]{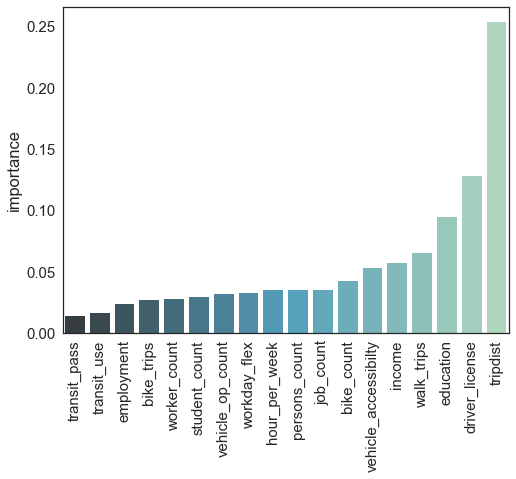}
    \caption{Mode prediction problem}
    \end{subfigure}
    \hfill
    \begin{subfigure}{.5\linewidth}
    \includegraphics[width=1\textwidth]{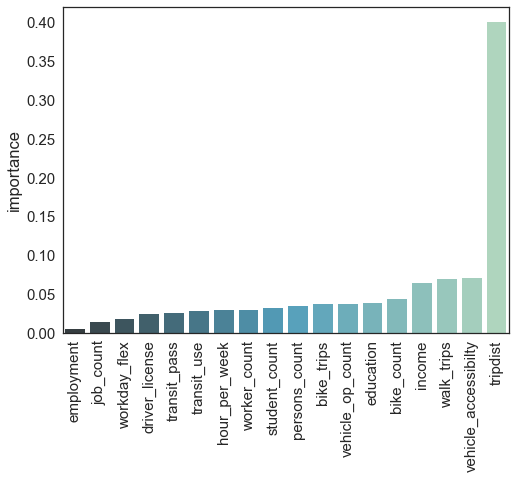}
    \caption{Category prediction problem}
    \end{subfigure}
    \caption{Gini importance \citep{breiman1984classification} of various features in predicting mode and category using the random forest model}
    \label{fig:importance}
\end{figure}

A deeper look (Figure \ref{fig:importance}) into the random forest classifiers reveals the Gini importance \citep{breiman1984classification} of various features in predicting the mode and category. As expected, we can see that the distance of the trip is the best predictor. People prefer to drive when they have to travel longer distances and there is no reliable public transport option. Apart from trip distance, characteristics of a person's transportation network (such as accessibility of vehicles in their home) and their past experience (such as walking trips undertaken last week) also contain predictive information. 

\subsection{Towards Computing Acceptability} 
For our candidate definitions of acceptability (in Section \ref{sec:defining-acceptability}), $P(r,p)$ and $P(u,p)$ are important quantities. In this section, we demonstrate how these quantities can be estimated from a population survey dataset. In particular, a person $p$ is defined as a collection of attributes encoding their demographic and employment information. Our ML approach is a significant step forward from previous empirical approaches in transportation research \citep{ben1985discrete} that have relied on standard features such as distance, income, etc. to predict mode selection. We conducted detailed interviews and survey \citep{mohan2019exploring} and found that a large variety of factors play a role in mode selection. Our ML-based approach here allows for a breadth of features including trip features (trip distance) as well as features that represent a person's transportation network (accessibility of various modes) and their experience (walking trips in the past week).  It should be noted that while these previous approaches are more detailed in measuring the relative importance of each attribute of an individual's preference, our approach is extensible to whatever attributes are available. As we discussed in Section \ref{sec:defining-acceptability}, for a planning formulation that seeks to select from various available plans, detailed models are not necessary. An approach such as ours that provides an overall evaluation will suffice.

Computing mode acceptability using these likelihoods captures the intuition that it is easier to switch to modes taken by similar people in similar contexts. Thus, our primary assumption is that the utility function (and consequently, acceptability) can be generalized to people with similar characteristics. For example, a traveller who usually drives to a destination lives in a neighborhood where 50\% people drive, 40\% use transit, and 10\% walk to the same destination. Intuitively, it should be easier for the traveller to switch to transit than to walk.



\section{Evaluating Acceptability: Evidence from a Stated Preference Study}
\label{sec:evaluation_acceptability}
Recall from Section \ref{sec:defining-acceptability} that not only do we have three candidate definitions of acceptability, it has not been previously shown that acceptability of a mode (computed using the ML models) will in fact impact adoption. Further, it is unclear if individuals differ in how acceptability of a recommendation impacts its adoption. These differences may arise due to difference in demographics (i.e., young individuals are more open to trying newer ways to travel) or due to some aspects of their personality that make them more or less susceptible to persuasion \citep{cialdini2001harnessing}. In this section, we study these questions. We rely on stated preference methods \citep{johnston2017contemporary} that are used to elicit customer preference as well as for product valuation via survey responses. In particular, we conducted a small choice experiment for mode adoption. The survey instrument was adapted for our use-case of recommendations made through a travelling assistant application. This experiment had the following goals:
\begin{enumerate}
\item to establish that when recommendations are provided to regular drivers, they will consider adopting the recommended mode;
\item to evaluate if adoption of the recommended mode depends on its acceptability computed via the ML models in Section \ref{sec:ml_approach};
\item to study individual factors in the impact of acceptability on mode adoption; and finally,
\item to choose a specific definition of acceptability and develop a working model of adoption to integrate with planning. 
\end{enumerate}

\begin{figure}[t]
    \centering
    \begin{subfigure}[b]{0.6\textwidth}
        \includegraphics[width=\textwidth]{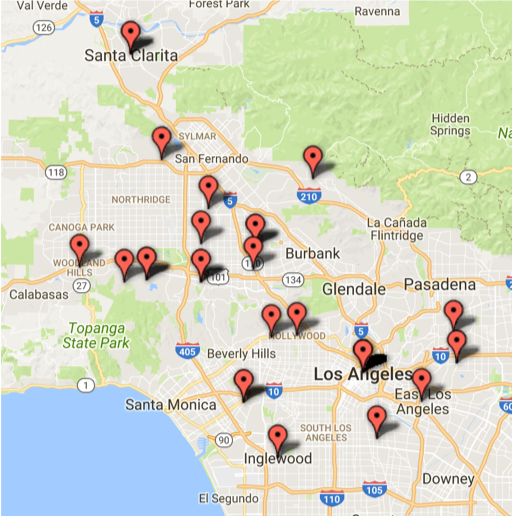}
    \end{subfigure}
    \hfill
    \begin{subfigure}[b]{0.27\textwidth}
        \includegraphics[width=\textwidth]{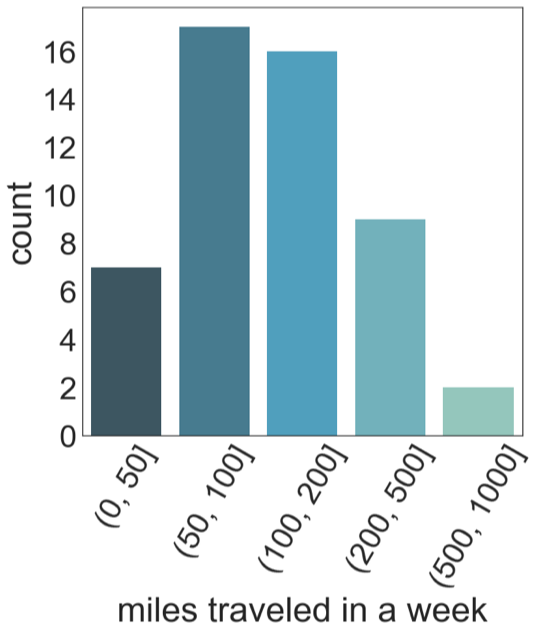}
        \includegraphics[width=\textwidth]{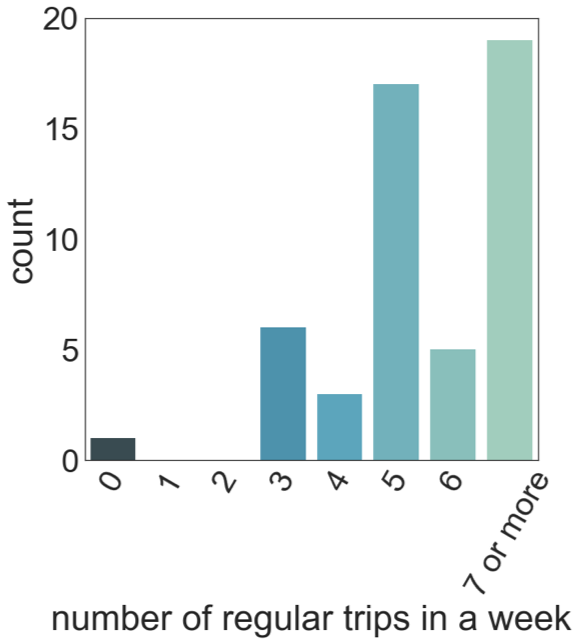}
    \end{subfigure}
    \caption{Location of study participants in the Los Angeles Metropolitan Area and their weekly travel characteristics.}
    \label{fig:stated-pref-participants}
\end{figure}

A choice experiment presents a participant with a set of options that differ on various attributes and asks the participant to select one option. This experiment and the following analysis uncovers how participants value different features. In our context of mode recommendation, one of the options is traveling by driving - the \emph{usual} mode - and another option is the \emph{recommended} mode determined by \textsc{Copter} through acceptable planning. We designed a choice experiment that mimics our context as described next. Our experiment had $49$ ($27$ female, $22$ male) participants who are regular drivers in the Los Angeles area. The participants had a mean age of $38.98(\pm 12.25)$ and a median household income of $\$75,000 - \$99,999$. The participants were distributed in the Los Angeles area and have diverse commuting behaviors as shown in Figure \ref{fig:stated-pref-participants}.
The participants were gathered by a paid recruiter who ensured compliance with responses. The participants were provided Amazon gift cards worth $\$30$ for completing the study. 

\subsection{Materials} 
Each study participant took two surveys within a week: 
\begin{enumerate}
\item \emph{Profiler}: For each participant, this survey collected a set of trips they regularly made by driving in terms of the origin, destination, departure time, and purpose. Participants were asked to provide a specific name for each trip (such as \emph{Film Networking Events} in Figure \ref{fig:question}). This name was used in a later survey to refer back to this trip. Additionally, we measured participants on the features incorporated in our random forest classifier (from Section \ref{sec:ml_approach}). To study individual differences in the impact of acceptability on adoption, we collected information about participant demographics: their age and gender as well as their susceptibility to persuasion using Cialdini's scale \citep{cialdini2001harnessing}. The $17$ item scale measures a participant's agreement with various values that underlie persuasion and influence: \emph{reciprocity} ($2$ items), \emph{commitment} ($2$ items), \emph{social proof} ($2$ items), \emph{authority} ($2$ items), \emph{liking} ($2$ items), and \emph{unity} ($5$ items). Measurements are done on a $7$-point Likert agreement scale from \emph{strong disagree} to \emph{strongly agree}. 

\item \emph{Mode adoption}: The mode adoption experiment was generated programmatically from the data collected in the profiler survey. We generated $10$ mode adoption questions for each participant. For all trips reported in the profiler survey, we generated a set of available modes. This set included: \emph{public transit}, \emph{driving slow}, \emph{driving 15 minutes later}, \emph{biking} if the trip distance is $<3$ miles and the participant owns a bicycle, and \emph{walking} if the trip distance is $<1$ mile. Each available mode must correspond to a class from the mode prediction problem in the previous section. We included driving-related modes as a sanity check for the responses in study. Because both driving modes ask participants to change their behavior, and consequently, will result in a negative switching gain. We expect that due to this negative switching gain, they will not be fully adopted. Further, because both are driving-related (and are similar to what the participants are already doing), their adoption will be better than taking the public transport or walking. Evidence suggests that the driving-related modes can reduce transportation energy consumption. From the trip-mode pairs, we randomly selected $10$ pairs. For each trip-mode pair, we generated a recommendation to be presented to the participant. This recommendation (see  Figure \ref{fig:question}) included information about the original trip: name of the trip, origin and destination addresses, day of the week, usual departure time, and usual arrival time; and the new recommended trip: recommended mode, departure time, and expected arrival time. Additionally, a visual representation of the proposed route was shown. The visual representation was generated using a web application - \textsc{TripGo} \cite{tripgo}. The participants were provided the context of this recommendation and asked to evaluate if they will take the recommendation. 

\begin{figure}[h]
  \centering
    \includegraphics[width=1\textwidth]{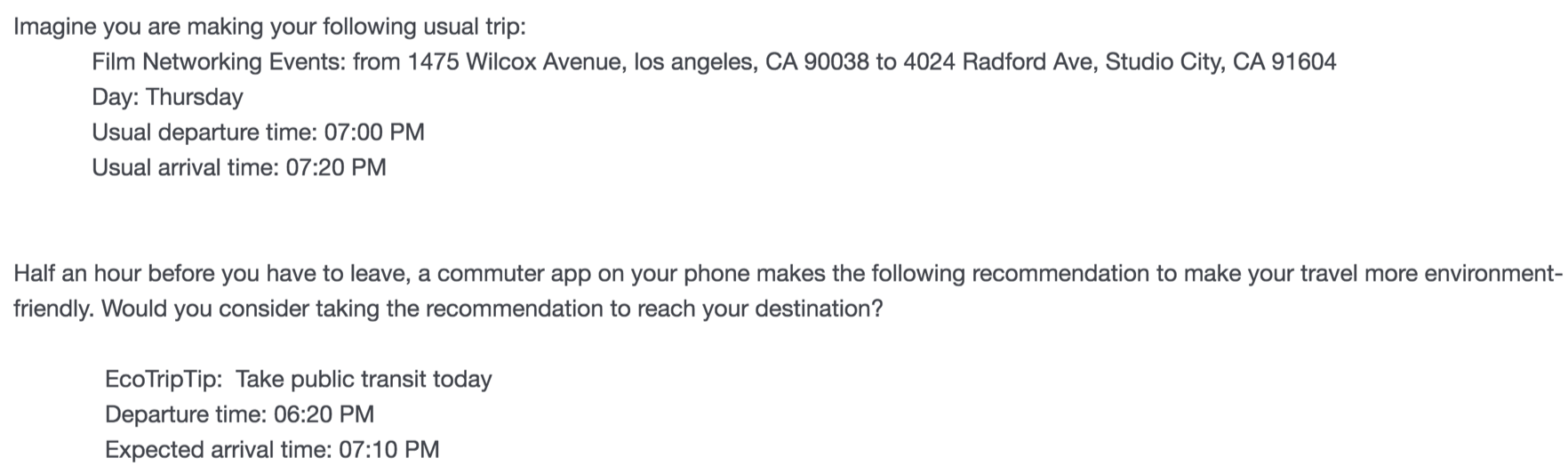}
    \includegraphics[width=1\textwidth]{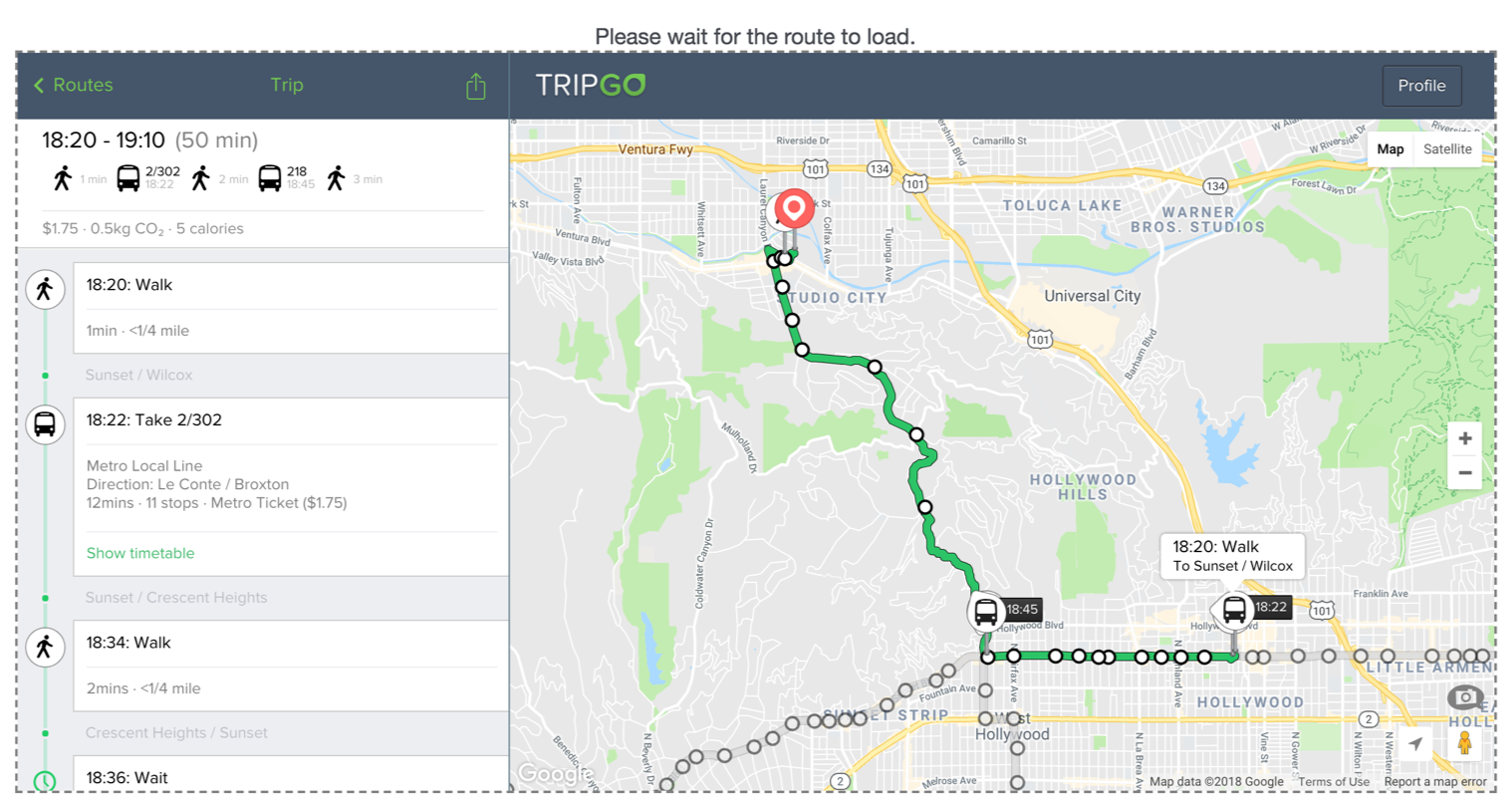}
    \includegraphics[width=1\textwidth]{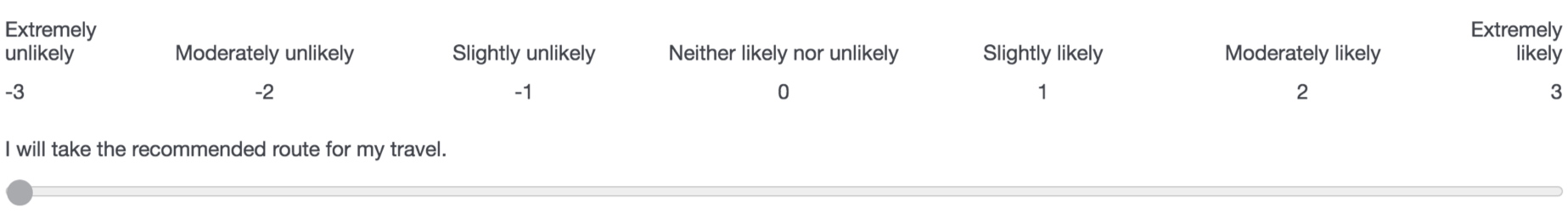}
    \caption{An example of the mode adoption question in the survey along with the response measurement instrument.}
    \label{fig:question}
\end{figure}

This question sets up the choice experiment. Participants had indicated that their usual mode of travel for that trip is driving. This recommendation sets up an alternate scenario and the participants are asked if they are willing to change their behavior by accepting the recommendation.  The participants indicated likelihood of accepting the recommendation using a $7$-point Likert scale: from \textit{extremely unlikely} ($-3$) to \textit{extremely likely} ($3$). If the participant is tending towards the \textit{unlikely} end of the scale, they can be considered choosing their usual mode, driving. When the participant is tending towards \textit{likely}, they can be considered choosing the recommended mode.
\end{enumerate}

\subsection{Data Analysis}
From the materials described above, we have the following for each mode adoption measurement:
\begin{enumerate}
    \item{Independent variables}
        \begin{enumerate}
            \item \emph{Computations for acceptability}. We estimated the probability of driving, taking public transit, and taking a non-motorized mode (cycling or walking) using the random forest ML model for category prediction described previously. Using these probabilities, we computed the following three values of acceptability as per the three candidate definitions of acceptability:  
            \begin{enumerate}
                \item switching gain: $\Delta_{r,u,p}$
                \item odds: $e^{\Delta_{r,u,p}} = Pr(r,p)/Pr(u,p)$
                \item likelihood of the recommended mode: $Pr(r,p)$ 
            \end{enumerate} 
            \item \emph{Individual variability}.
            \begin{enumerate}
                \item \emph{Demographics}, measurement of gender and age.
                \item \emph{Susceptibility to persuasion}, measurements on $17$ item Cialdini scale. 
            \end{enumerate}
        \end{enumerate}
    \item{Dependent response variables}
        \begin{enumerate}
            \item \emph{Ordinal adoption}: Participant response on the $7$-point Likert scale.
            \item \emph{Binary adoption}: To simulate real-world behavior where someone will either follow the recommendation or not, we compressed the ordinal response by categorizing everything reported as \emph{slightly likely} or higher as $1$ and others at $0$.
        \end{enumerate}
  
\end{enumerate}

Using the data above, we conducted the following analyses:
\begin{enumerate}
    \item In Section \ref{sec:driving_adoption}, we plot the distribution of ordinal adoption responses for various modes to show that participants consider adopting modes when provided with a recommendation and that our instrument - the mode adoption survey - is a reasonable instrument. 
    \item In Section \ref{sec:acceptability_selection}, we evaluate if acceptability of a mode (computed using population-based ML models) is predictive of adoption and which candidate definition should be adopted. To do this, we employ mixed-effect linear and logistic regression models for the ordinal and binary adoption respectively. Mixed-effects linear regression is of the form $y = \alpha + \beta x + \gamma z + \epsilon$, where $y$ is the adoption vector, $x$ is a fixed-effect vector corresponding to different definitions of acceptability, and $z$ is a random effect vector corresponding to a participant. $\alpha$ is the intercept, $\beta$ and $\gamma$ are coefficients, and $\epsilon$ is the error term. For the binary adoption variable, we fit a mixed-effects logistic regression of the form $Pr(y) = 1 / ( 1 + e^{-(\alpha + \beta x + \gamma z + \epsilon)})$, where $Pr(y)$ is the probability of observing the specific adoption value ($0$/$1$). Participants are modeled as random effects in these models under the \emph{random intercept} assumption which encodes that some variance in responses is due to each participant having a different baseline reaction to recommendations. We report $R^2m$, the proportion of variance in $y$ explained by measured, independent variables as well as $R^2c$, the proportion of the variance that is explained by the complete model including the random individual-level effects.
    \item In Section \ref{sec:adoption_individual}, we explore the individual differences in the impact of acceptability in adoption of recommendations. Here, we include participants' measurements on the Cialdini's scale as fixed-effects in the regression models as well as their age and gender. 
    \item In Section \ref{sec:adoption_model}, we select a definition of acceptability and develop a working model of adoption for formulation of acceptable planning. 
\end{enumerate}

\subsection{Results}
\subsubsection{Adoption of Driving v/s Other Mode Recommendations}
\label{sec:driving_adoption}
First, we compare the adoption rates of driving-related modes with others. As discussed earlier, \emph{H1}: we expect the adoption rates of driving-related modes to be higher than other modes as they are similar to what the participants are already doing. However, \emph{H2}: we expect them to not be fully adopted as they do require a change in usual behavior.

\begin{figure}[h]
    \centering
    \includegraphics[width=0.6\textwidth]{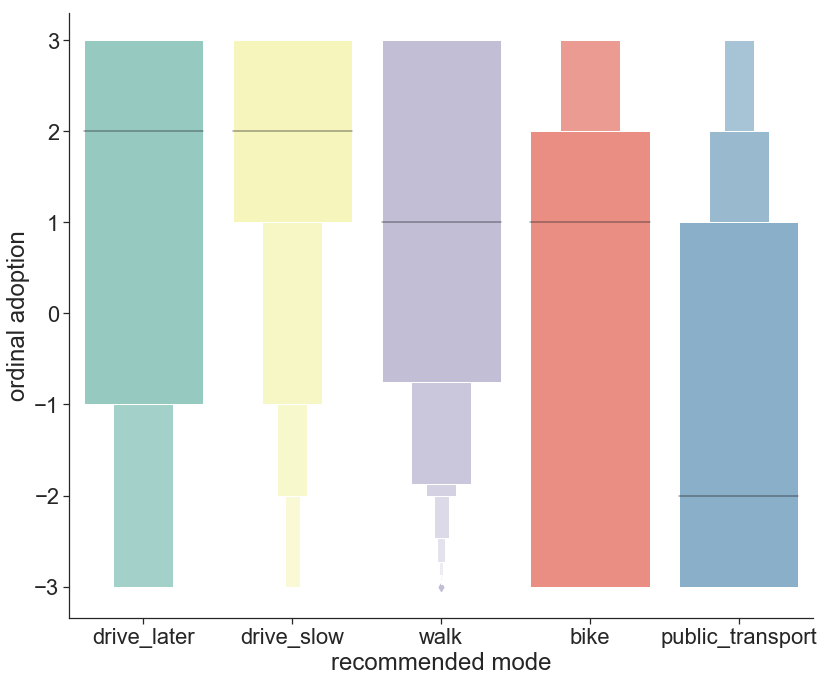}
    \caption{Box plots showing the distribution of ordinal adoption responses by mode.}
    \label{fig:box-plots-modes}
\end{figure}

We compare ordinal adoption of various recommended modes by plotting Likert-scale ratings as in Figure \ref{fig:box-plots-modes}. We see that both \emph{H1} and \emph{H2} are supported in our data. Further, we find that adoption of \emph{driving slow} may be better than that of \emph{driving later}. That is reasonable, given that for some participants arriving at their destination by a certain time may be required. Given that the data collected through the choice experiment study matches our intuition about the domain, we can conclude that the study is measuring useful information about participants' behavior.

\subsubsection{Acceptability and Mode Adoption}
\label{sec:acceptability_selection}
Next, we study different definitions of acceptability proposed in Section \ref{sec:defining-acceptability}. Recall that we have three candidate definitions:
\begin{enumerate}
\item switching gain, $\Delta_{r,u,p}$
\item odds, $e^{\Delta_{r,u,p}} = Pr(r,p)/Pr(u,p)$ 
\item probability, $Pr(r,p)$ 
\end{enumerate}
where $r$ is the recommended mode, $u$ is the usual mode for person $p$ and $Pr()$ is the likelihood of prediction from our ML models. Results from mixed-effect linear and logistic regression are shown in Table \ref{tab:regression}. For each definition of acceptability, we report $R^2c$, the proportion of the variance that is explained by the complete model, and $R^2m$, the proportion is explained by the candidate definition. Each cell in the table denotes a fitted mixed-effects regression model.

Overall, the results show that people reported a willingness to change their travel behavior by adopting recommended modes. Further, we see that all definitions of acceptability significantly impact the adoption of a recommendation. The coefficients align with the intuition about the domain - as the cost of adopting the recommendation decreases (acceptability increases), adoption increases. We see that $R^2c$, $R^2m$ are highest for odds. These results suggest that it is the best predictor of mode adoption because it explains the most variance in the data. The odds and probability have similar impact on adoption due to the fact that our participants are regular drivers and $Pr(u,p) \approx 1$ for many of the trips. With a more diverse population in a larger deployment, the differences between these definitions of acceptability can be teased apart. For further analyses and implementation, we chose the odds definition of mode acceptability.

    \begin{table}
        \centering
        \small
        \def\arraystretch{1.2}%
        \begin{tabular}{lll}
        \toprule
            Dependent $\rightarrow$ & \multicolumn{1}{c}{Adoption} & Adoption \\
            Independent $\downarrow$ & \multicolumn{1}{c}{(ordinal)} & \multicolumn{1}{c}{(binary)} \\ \midrule
            (intercept) &-0.017&-0.185\\
            switching gain, $\Delta_{r,u,p}$&0.108*&0.104*\\  \midrule[0.15pt]
            $R^2m$&0.034&0.035\\
            $R^2c$&0.347&0.270\\ \midrule
            (intercept) & -0.025 & -1.065 \\
            odds, $e^{\Delta_{r,u,p}}$ & 2.386***  & 1.780** \\ \midrule[0.15pt]
            $R^2m$ & 0.075 & 0.064 \\
            $R^2c$ & 0.379 & 0.346 \\ \midrule
           (intercept)&-0.964&-1.080\\
            probability, $Pr(r,p)$&3.623***&3.317*\\ \midrule[0.15pt]
            $R^2$m &0.066&0.058\\
            $R^2$c &0.369&0.293\\ \bottomrule
            \end{tabular}
        \caption{Regression modeling results for ordinal and binary mode adoption. *** p $< 0.001$, ** p $< 0.05$, * p $< 0.1$}
        \label{tab:regression}
    \end{table}
    
    \begin{figure}
        \centering
        \includegraphics[width=0.45\textwidth]{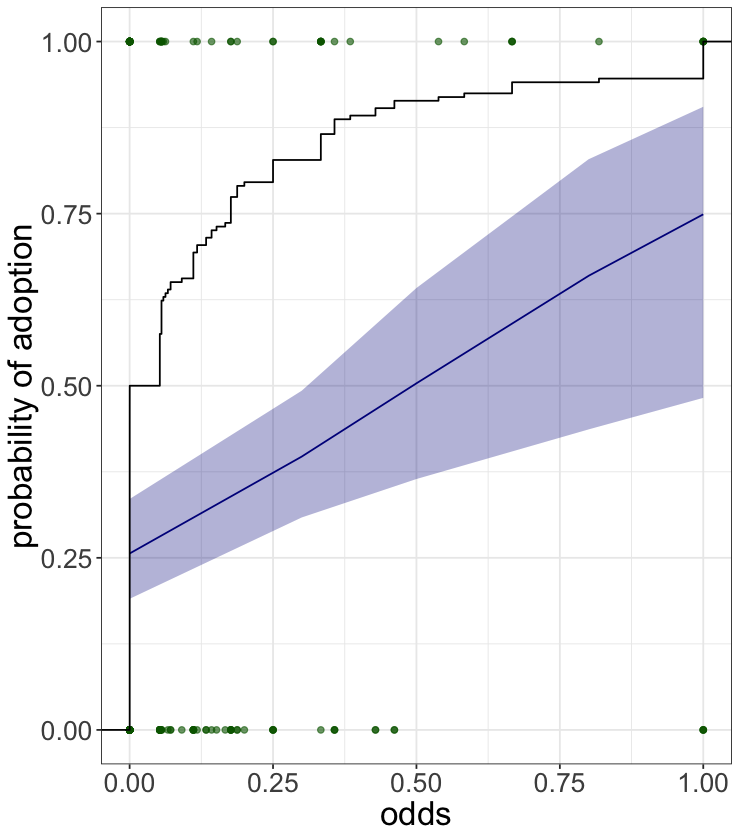}
        \caption{Participant mode adoption (green), cumulative density of responses (black), and results of the mixed-effects logit model with odds as independent variable (blue)}
        \label{fig:logit-model}
    \end{figure}

Figure \ref{fig:logit-model} shows the distribution of responses and the results of the regression model with odds as the independent variable. The green points represent adoption choices by different participants. The black line shows the cumulative density of responses. The blue line and ribbon capture the coefficient of odds and the confidence interval.  From the graph, it can be seen that the statistical significance arises from the responses where the estimated likelihood ratio $> 0.5$ ($<= 10\%$ of responses). For lower values of likelihood ratio, the responses are evenly distributed. One explanation of these observations is that when asked about changing behavior to support sustainable modes of transport, participants tended to be optimistic. In a deployment where real choice data is available, revealed preference methods \citep{carson1996contingent} can be employed for more accurate choice modeling.

\subsubsection{Individual Variability in Adoption}
\label{sec:adoption_individual}
Results in the previous section provide strong evidence that acceptability of a mode is a significant predictor of its adoption. Next, we study if there are individual differences in adopting a mode recommendation above and beyond its computed acceptability. We have two kinds of factors that may bring individual variability in adoption: demographics and personality - in particular, susceptibility to persuasion. We analyze the impact of these factors by extending the regression models to include these factors as independent variables in two different efforts. To be valuable, these factors must influence adoption significantly as well as increase the variance explained by fixed-effects of the model ($R^2_m$). 

First, in Table \ref{tab:regression_age_gender}, we show results from the regressions that include an ordinal variable representing participant age and a categorical gender variable. As is apparent, there is no significant impact of either age or gender, suggesting that these variables do not play a role in adoption. 

\begin{table}[h]
\small
\centering
\def\arraystretch{1.2}
\begin{tabular}{lll}
\toprule
    Dependent $\rightarrow$ & \multicolumn{1}{c}{Adoption} & Adoption \\
    Independent $\downarrow$ & \multicolumn{1}{c}{(ordinal)} & \multicolumn{1}{c}{(binary)} \\ \midrule
    (intercept) & -1.63736 & -1.67021 \\
    odds, $e^{\Delta_{r,u,p}}$& 2.38833*** &  2.15086** \\ 
    GenderMale & 0.33149 & 0.01281 \\
    Age & 0.01377 & 0.21983 \\ \midrule[0.15pt]
    $R^2m$ & 0.085 & 0.072 \\
    $R^2c$ & 0.396 & 0.310 \\ 
\bottomrule
\end{tabular}
\caption{Regression models with age and gender for ordinal and binary mode adoption. *** p $< 0.001$, ** p $< 0.05$, * p $< 0.1$ \newline \newline}
\label{tab:regression_age_gender}
\end{table}

Next, we extended the mixed-effects regression model to include independent variables from the susceptibility to persuasion scale \citep{cialdini2001harnessing}. In Table \ref{tab:regression-full}, we show the results from these regressions. As previously, we model both ordinal and binary adoptions, shown as models of class I and II in the table. For each class, we develop three models with varying independent variables. Models IA \& IIA are reproduced from Table \ref{tab:regression} for comparison. Models IB \& and IIB are full models and include all measurements from the Cialdini scale as independent variables. Models IC \& and IIC are simplified models that only have variables judged to be significant in the full models.

In models IB, we see that commitment (i7) and liking (i12) dimensions are significant predictors of adoption. As is expected, $R^2_m$ and $R^2_c$ both increase significantly suggesting that including these variables in the model helps explain more variance in our dataset, above and beyond what is explained by odds. In model IC, in which we removed non-significant persuasion items, we see that both commitment and liking dimensions are still significant. Further, we see that $R^2_c$ in models IC is close to that of model IA suggesting that they explain similar amounts of variance. However, in model IC $R^2_m$ (variance explained by fixed-effects) is much higher than that of model IA. A proportion of variance that was explained by random effects (corresponding to participants) in model IA can be explained by measured persuasion items in model IC. Similar observations can be made of Models II. This suggests that persuasion items measure important predictive information about individual adoption of recommendations above and beyond what odds does. 

\begin{table}[t]
        \footnotesize
        \centering
        \def\arraystretch{1.1}%
        \begin{tabular}{lccccccc}
            \toprule
            \textbf{MODEL} & \multicolumn{3}{c}{I} & \multicolumn{3}{c}{II}\\
        & \textbf{A} & \textbf{B} & \textbf{C} & \textbf{A} & \textbf{B} & \textbf{C}\\
        \textbf{Dependent $\rightarrow$} & \multicolumn{3}{c}{Ordinal adoption} & \multicolumn{3}{c}{Binary adoption}\\
        \textbf{Independent}$\downarrow$ &\multicolumn{3}{c}{}&\multicolumn{3}{c}{}\\
        \midrule
        (intercept) &-0.025&-0.183 &-0.201 & -1.065&-0.084& -0.998 \\
        odds, $e^{\Delta_{r,u,p}}$&2.386***& 2.306*** & 2.348***&2.1587**&1.780** &1.729**\\
        I1 \emph{Reciprocity repay favor}&&   -0.138&&& -0.166&\\
        I2 \emph{Reciprocity return gift}&&   -0.009&&& -0.043&\\
        I3 \emph{Scarcity last to buy}&& -0.034&&& -0.009&\\  
        I4 \emph{Scarcity special value}&&  -0.212&&&-0.170&\\    
        I5 \emph{Authority obey directions}&&   0.311&&&0.330*&0.1386\\
        I6 \emph{Authority listen to}&&   -0.002&&&0.021&\\     
        I7 \emph{Commitment appointment}&& 0.495*&0.515**&&1.126**&0.979***\\  
        I8 \emph{Commitment promise}&&  0.448&&& 0.207 &\\
        I9  \emph{Consensus advice}&&   -0.051&&& 0.031&\\   
        I10 \emph{Consensus what to do}&&   0.170&&& 0.206&\\ 
        I11 \emph{Liking inclined to believe}&&  -0.069&&& 0.075&\\    
        I12 \emph{Liking do favor}&&  -0.966**&-0.819**&&-1.105**&-0.905**\\  
        I13 \emph{Unity help}&& -0.143&&& -0.238&\\ 
        I14 \emph{Unity self assembly furniture}&&  -0.026&&& -0.008&\\    
        I15 \emph{Unity country}&&   0.236 &&& 0.128&\\  
        I16 \emph{Unity sports team}&&   0.157 &&& 0.162&\\ 
        I17 \emph{Unity vicarious success happy}&&  -0.357 &&& -0.491&\\
        \midrule
        $R^2m$ &0.075&0.224&0.156&0.064&0.329&0.254\\
        $R^2c$ &0.379& 0.467&0.388&0.301&0.346&0.318\\
        \bottomrule
        \end{tabular}
        \caption{Regression modeling results for ordinal and binary mode adoption. *** p $< 0.001$, ** p $< 0.05$, * p $< 0.1$}
        \label{tab:regression-full}
\end{table}

\subsubsection{Working Model of Adoption}
\label{sec:adoption_model}
In this section, we first demonstrated that providing recommendations to people in a travel route choice task can indeed influence their decisions. Then, we evaluated if acceptability (odds, as defined in Section \ref{sec:defining-acceptability} and computed using ML model in Section \ref{sec:ml_approach}) impacts adoption. Our analysis suggests that it indeed is a significant factor in recommendation adoption. Further, we show that different individuals may be influenced to a different extent. In our analysis, a person's measurement of susceptibility to persuasion on Cialdini's scale is useful in explaining this individual variation in the impact of acceptability on adoption.

To identify the recommendation that reduces the most expected energy, it is necessary to estimate the likelihood of adoption. Recall that our motivating example suggests that a recommendation should be \textit{timely}, \textit{acceptable}, and \textit{compelling} among other things to motivate behavior change. At this point, we have defined, computed, and evaluated what acceptability is. Therefore for the remainder of this paper, as a working model of adoption, we use the logit model of odds to estimate the probability of a person adopting a recommended route. Understandably, this is a limited model of adoption and our future efforts will extend this model further to include other factors as well. In particular, our results in Section \ref{sec:adoption_individual} suggest that the framing of a recommendation can be personalized to every individual such that it is compelling to them. We will extend our adoption model to include the impact of personalized compelling framing. 

\section{Finding Energy-Efficient Acceptable Plans}
\label{sec:acceptable_planning}
Until now, we have defined plan acceptability, proposed a data-driven way to compute it, and have validated it to be an important component of mode adoption. Now, we extend the multi-modal planning formulation to include acceptability. The main concern in incorporating acceptability in planning is that costs in planning are defined at edges, while acceptability is an overall evaluation of a plan recommendation. For integration, we sacrifice efficiency and split plan evaluation in two phases as described below.

\begin{figure}[h]
  
  \centering
    \includegraphics[width=1\textwidth]{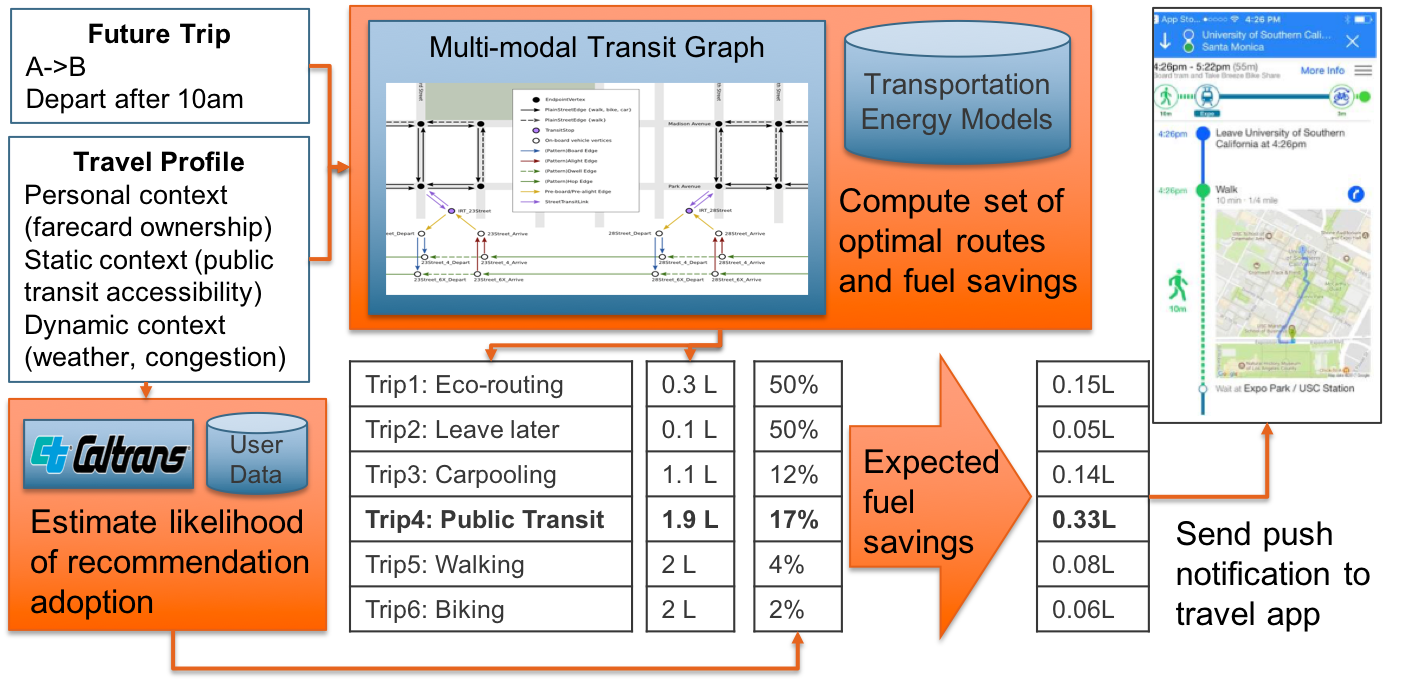}
    
    \caption{\textsc{Copter} identifies the recommendation with the largest expected reduction in energy consumption by acceptable planning and energy modeling.}
    \label{fig:two-stage}
\end{figure}

Figure \ref{fig:two-stage} provides an overview of \textsc{Copter}'s operation. In a deployment, we expect to maintain a series of future trips for each user. Before an expected departure, \textsc{Copter} computes an energy efficient \emph{acceptable} alternative and sends it to the user $15$ minutes before they would have to leave for the new trip.  To compute this plan, we augment the multi-modal planning formulation with our acceptability model with the following steps:
\begin{enumerate}
\item Generate a mode candidate set $M_p$ for the person $p$ for whom the request is made. $M_p$ can be generated by knowing if they can walk and if they have a bike. For example, someone who doesn't own a bike but can walk $M = \{walk (w), bus (b), subway (s)\}$.
\item Given $M_p$, determine the language $L_p$ set that is valid. For the example above, $L_p = \{w* , w*b+w*, w*s+w*\}$. Note that while this is a fairly simplistic set, the formulation can be extended to include more complex plans. 
\item For every element in $L_p$, generate the most time-efficient plan (using the multi-modal formulation) where $cost(e) = dur(e, m)$ where $m$ is the mode used to traverse the edge. This process will generate a candidate set of plans for the person, $\Pi_p$. 
\item Compute the energy reduction in each plan $\pi \in \Pi_p$ using an existing mesoscopic energy model \citep{elbery2018}.
\item Evaluate the likelihood of adoption, $adopt(\pi, p)$, for every $\pi \in \Pi_p$ using the logit model of adoption in the previous section. 
\item \label{item:plan-selection} Select a plan that has maximal expected energy savings, $\pi^* = \argmax_{\pi \in \Pi_p} adopt(\pi, p) \times energy(\pi)$. 
\end{enumerate}

There are two items in the above process that merit additional discussion. First, is the proposed acceptable planning framework extensible to additional multi-modal route recommendations? For each element in the candidate mode set $M_p$, there must be a language $L_p$ that constrains the possible sequences of modes and corresponding likelihood of adoption, $adopt(\pi,p)$. In the empirical work discussed in this paper, the likelihood of adoption is the same for all trips involving public transit (an element of $M_p$). To consider trips where the traveler walks to transit distinctly from those in which the traveler takes a taxi for the first/last mile requires changing $M_p$ and $L_p$ as well as gathering mode adoption data to update $adopt(\pi,p)$ for the new modes in $M_p$. Computing $adopt(\pi,p)$ requires the acceptability of the plan which in turn is estimated from travel behaviors of a population. As data collection in cities gets more sophisticated, we expect to have a richer dataset of mode sequences for trips. This enables the use of classification-based likelihood estimation to compute adoption.

Second, to clarify the plan selection (Step \ref{item:plan-selection}), consider two options for a particular traveler: walking which saves huge amounts of energy but its likelihood of adoption is very low and taking public transit which does not save as much energy but has much higher likelihood of adoption. Based on quantitative expected savings in fuel, \textsc{Copter} may recommend walking by ranking it above public transit, which may not result in adoption and consequently, no savings in fuel. If plan selection is biased by likelihood of adoption alone, \textsc{Copter} will recommend public transit which has improved chances of adoption and consequently, is more likely to result in some energy savings. From a behavior change perspective, the second approach may be better - some change is better than none at all. This is a design question that can be answered empirically by deploying both competing plan selection approaches. However, in absence of empirical observations from deployments, we consider the algorithm above for further evaluation. Our formulation allows for exploring both design criterion in future.

\section{Energy Simulation Experiments}
\label{sec:experiments}
\begin{figure}
  \centering
    \includegraphics[width=0.5\textwidth]{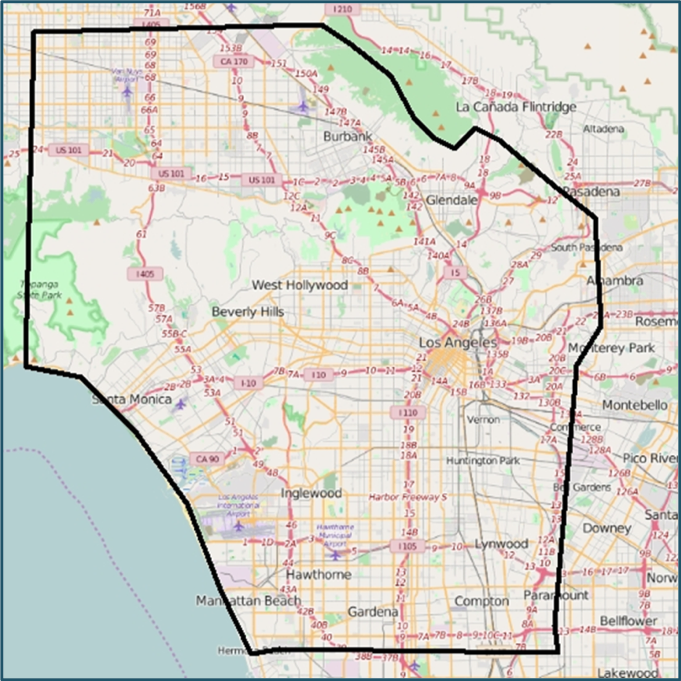}
     \caption{Los Angeles region study area.}
       \label{fig:study-area}
\end{figure}

While deploying \textsc{Copter} in a large city to measure the impact is the ideal evaluation, it is extremely resource intensive to conduct. Furthermore, without being incorporated into a widely used product, it would be impossible to iteratively evaluate different components of a complex system such as \textsc{Copter} employs. Therefore, we evaluate the potential impact of \textsc{Copter} using a high fidelity simulation of the Los Angeles region as shown in Figure \ref{fig:study-area}. Our integrative simulation simulates people of Los Angeles making commuter trips in the morning and evening periods. It uses the adoption model developed in the previous sections to simulate \textsc{Copter}'s influence on their decisions and computes energy impact using a state-of-art transportation energy simulation of the LA region described briefly below.

\subsection{Transportation Energy Model}
The simulation model \citep{elbery2018} supports city-level networks with different modes of transportation including cars, buses, railways, walking, biking, and carpooling. It utilizes both microscopic and mesoscopic simulation to leverage their respective strengths of accuracy and scalability. The simulation spatially partitions the road network enabling distinct portions of the region to be simulated in parallel.  Additional parallel simulations capture loosely interacting modes (e.g., the rail and road network). This enables micro-simulation (i.e., considering driver behavior w.r.t. throttle position, braking, and lane choice every $0.1$s) of the arterial and highway roads encoded as 10,650 links.

Dynamic calibration aligns previous government planning documents with observed vehicle flows to capture the changing rates of vehicles entering the different networks throughout the day \citep{Du2018}. The simulation incorporates previously validated driver behavior \citep{rakha2009simplified} and fuel \citep{fiori2016power} models to provide accurate estimates of travel time and energy consumption due to traveler influence. 

\subsection{Simulation Scenario}
Our study area, shown in Figure \ref{fig:study-area}, covers the majority of Los Angeles County. Our study area includes diverse offerings with over $170,000$ roadway links and over $1$ million daily transit trips.  During the AM period ($7$am-$10$am) and PM period ($4$pm-$7$pm), our calibrated transportation model includes $1.2$ and $1.7$ vehicle trips respectively.  As the effects of congestion continue past the peak period, our study considers an AM peak/off-peak period ($7$am-$12$pm) and a PM peak/off-peak period ($4$pm-$9$pm).

In our deployment scenario, we assume that $10\%$ of peak period drivers use apps (e.g., Waze, TripGo) through which we can offer alternative suggestions.

\subsection{Simulation Method}
Given our deployment assumption that 10\% of the driving population receives our messages, we sample $10\%$ of the traffic during the peak periods ($7$am-$10$am and $4$pm-$7$pm) for our influenced population.  Our experiment has two conditions:
\begin{enumerate}
\item Baseline: Each influenced traveler drives their original route
\item Influence: Each influenced traveler probabilistically takes the maximum expected energy reduction route as determined by \textsc{Copter} using the mesoscopic model or drives their original route based on working model of adoption in Section \ref{sec:adoption_model}. 
\end{enumerate}

To account for the uncertainty in alternative adoption and background traffic rates, we ran each condition four to seven times. The simulation model runs in just over real time and therefore, running a larger number of trials is prohibitive. $5-10$ trials is standard practice for similar transportation studies. We report $95\%$ confidence intervals for difference of means between the Baseline and Influence trials for total fuel consumption and congestion related delay.  We also report the resulting changes in mode of the influenced travelers.

For simulation in the influence condition, we:
\begin{enumerate}
    \item Sampled features from the CHTS dataset and used these features to augment trips in our energy models
    \item For each trip, selected a recommendation from motorized, non-motorized, and public transit using \textsc{Copter}
    \item Computed the likelihood of adoption by applying our working model of adoption as follows:
        \begin{enumerate}
            \item Generated an intercept by fitting and sampling from a Gaussian distribution with intercept mean and standard deviation from our analysis.
            \item Computed acceptance likelihood by applying the working model of adoption
        \end{enumerate}
    \item Simulated adoption by selecting the recommendation biased by likelihood of adoption
\end{enumerate}

\subsection{Results}
Table \ref{tab:ampm} contain the results for AM and PM periods obtained using the microscopic model.  These results indicate a potential $4.6\%$ energy reduction during the AM period and $3.5\%$ energy reduction in the PM period.  Congestion-induced delay results were even greater at $20\%$ and $14\%$, respectively.  This is significant considering the influenced population represents about $6\%$ of the total traffic over the period.  To identify an upper bound, we ran simulations in which every influenced traveler walked.  In these simulations, we observed a reduction in energy consumption of $~9\%$ and a reduction in delay of $~30\%$ across the AM and PM periods. Table \ref{tab:mode} indicates that almost half the influenced population is willing to take alternative modes. These mode switches are based on our mode adoption study and may be optimistic. Future work will refine these estimates through a deployment of the technology and measuring people's behavior.

\begin{table}[h]
\centering
\def\arraystretch{1.2}%
\begin{subtable}{1\linewidth}
\centering
\begin{tabular}{lccc}
\toprule
 & Baseline & Influence & Change (CI)\\
 \midrule
Total& 3,195,637  & 3,048,278 & -4.6\%\\
Fuel (l)&&&(-3.6\% -5.6\%)\\
Total& 249,221 & 199,395 & -20\%\\
Delay (hr) &&& (-13.6\% -26.4\%)\\
\bottomrule
\end{tabular}
\caption{Results of departures in the AM period: 7:00 - 12:00}
\end{subtable}
\begin{subtable}{1\linewidth}
\centering
\begin{tabular}{lccc}
\toprule
 & Baseline & Influence & Change (CI)\\
 \midrule
Total& 3,487,982  & 3,367,675 &-3.5\%\\
Fuel (l)&&&(-2.6\% -4.3\%) \\
Total& 375,137 & 322,228 &-14.1\\
Delay (hr)&&&(-10\% -18\%)\\
\bottomrule

\end{tabular}
\caption{Results of departures in the PM period: 16:00 - 21:00}
\end{subtable}

\caption{The mean for each condition is reported along with the percent change in the Influence condition. We also report a $95\%$ confidence interval for the difference in mean.}
\label{tab:ampm}
\end{table}

\begin{table}[h]
\def\arraystretch{1.2}%

\centering
\begin{tabular}{lcc }
\toprule

Mode & AM Share & PM Share \\
\midrule
Car	& 54\% & 53\% \\
Walk & 42.7\% &	42.8\% \\
Bike & 3.6\% & 3.8\% \\
Bus	& 38.9\% & 39.1\% \\
Train &	14.4\% & 14\% \\
\bottomrule
\end{tabular}
\caption{Share of influenced population that used mode on trip.  Sum is over 100\% due to trips using multiple modes.}
\label{tab:mode}
\end{table}

\section{Related Work}
\label{sec:related_work}
This paper incorporates ideas from diverse disciplines and consequently, is related to a large body of work in transportation and AI. Here, we briefly discuss how our work relates to ongoing research. 
\subsection{Transportation}
The majority of transportation policy and social science research either focuses on systemic changes that are applied evenly to the entire population such as congestion pricing/dynamic tolling \citep{brownstone2003drivers} or studies long term behavior change in individuals to encourage sustainable commuting \citep{castellani2016understanding}. Recently, there has been a shift toward personalizing the interventions to increase the use of sustainable transportation. IncenTrip \citep{doi:10.1177/0361198118775875} allocates incentives from government sponsored programs such that each dollar spent provides the largest environmental impact. TriPod \citep{azevedo2018tripod} takes a different approach in which the allocation comes from a points-based system in which companies give travelers points for alternative transportation choices that can be redeemed at local businesses. While these systems primarily rely on monetary incentives to affect travel behavior, Metropia  \citep{zhu2018will} has shown that congestion information alone can influence traveler departure time. Our work differs from the above by identifying a single personalized recommendation from diverse alternatives based on network conditions without any monetary incentives. 

In addition to algorithmic work in route planning, the AI community has explored other aspects of the transportation influence problem. Work on compliant micro-tolling assumes a subset of agents would be willing to opt-in to a congestion tolling regime \citep{Hanna2019}. In that work, energy and traffic are modeled mesoscopically to compute marginal cost paths for social benefit and human behavior is captured using a single parameter for value of time. Recent work on recommender systems use more complicated user models to order potential routes to a user \citep{liu2019joint}. Our work differs by computing alternative routes to an expected default.

Other work in AI seeks to make predictions about the transportation network. Using position traces and other data, systems predict the mode and destination of travelers \citep{LIAO2007311,song2016deeptransport}. These works address travel prediction problems, which are essential to the commercial applicability of our approach. With respect to the transportation influence problem, \textsc{Copter} could incorporate these approaches as follows. First, predicting expected future trips is necessary to trigger the generation of alternatives. Second, the prediction of current mode supports monitoring travelers to determine if they follow the recommendation and to assist them as conditions change.

\subsection{AI for Social Good}
\citet{hager2017} identified urban computing and sustainability as one of the motivating domains for research on intelligent methods for public welfare. The methods developed in this paper take a unique perspective towards the problem of urban computing and sustainability in recognizing that humans are a critical component of urban transportation systems. It puts reasoning about human behavior as the center of intelligent system design. Our approach aligns with the perspective of \citet{abebe2018mechanism}, who argue that AI for social good inherently is an inter-disciplinary problem and developing good AI solutions requires deep grounding in the field of interest and collaboration with domain experts. To develop our AI planning-based approach, we engage with behavioral economics to understand how people choose and with transportation research to estimate energy consumption. 

Our work is a part of growing body of work in AI that seeks to achieve social good goals. Intelligent systems have been employed for allocation of security resources at ports \citep{shieh2012protect}, protecting biodiversity in conservation areas \citep{fang2016deploying}, and screening 800 million airport passengers annually throughout the USA \citep{brown2016one}. Research has also begun exploring intelligent algorithms and systems that can influence people's behavior to benefit society. Most of this work has been focused on health. \citet{hartzler2016acceptability, mohan2017designing} develop an adaptive interactive system for scaffolding learning of exercising behaviors in sedentary populations. \citet{yadav2018bridging} explore social network models to maximize influence for HIV awareness in homeless youth population. Our work brings technology-mediated influence and behavior change to the domain of urban computing and sustainability.

\subsection{Human-Aware AI Systems}
AI research on modeling of humans in human-AI collaborative systems is in its infancy. While the community has recognized the importance of incorporating explicit models of human behavior in AI systems to make them more effective in collaborative tasks \citep{khampapati2018, albrecht2018autonomous}, there is a dearth of computational models that can be incorporated with AI algorithms. In problems where the humans (or their teams) are adversaries to the AI algorithm, various game-theoretic modeling techniques \citep{sinha2018stackelberg} have been found to be useful. In dearth of detailed models of why a specific human behavior occurs, these approaches optimize against the worst case, providing guarantees of reasonable performance in the expected case. Often in collaborative modeling, task models employed by AI systems for acting are used as a stand-in for representing human goals and behaviors \citep{chakraborti2018projection,kim2015inferring}. These approaches are useful when humans and systems are engaged in the same task and have similar goals.

Our proposed method takes a unique perspective towards building models of human collaborators. In our problem, the human and the agent are engaged in different tasks. The human's goal is to reach their destination while the agent's goal is to perturb human cognitive system such that the human takes a sustainable mode to get there. We adopt the choice modeling techniques from behavioral economics and explore how it can be incorporated in the AI planning framework to produce personalized plan recommendations. In contrast with adversarial modeling techniques, our work incorporates explicit causal models of human decisions to produce effective solutions. Our work can be considered as an instance in the human-aware AI design agenda that exploits detailed, explicit models of human decisions, learning, and behavior and of which cognitive tutors \citep{Koedinger2013NewOptimization} are another instance. Along these lines, \citet{mao2012modeling} proposed human modeling for AI systems that is grounded in theory of human cognition and reasoning. It is worth noting that this work aims at only predicting human behavior while our work takes a step beyond prediction and employs predictive information for causing a change in behavior.

\section{Discussion and Conclusion}
\label{sec:conclusion}
This paper presents \textsc{Copter}, an intelligent travel assistant that is designed to influence individual travel behavior to reduce the total energy consumption of a large city. To do this, we propose an integrative framework that is a novel combination of multi-modal trip planning, transportation choice modeling, and machine learning. We introduce the notion of acceptability of a recommended alternative and define it in terms of the change in utility. We show how this change can be estimated using ML models trained on the CHTS dataset. We evaluate different definitions of acceptability using a human mode choice study. Finally, in a simulation experiment over the Los Angeles area we show that if $10\%$ of peak travelers were receiving influence messages, estimated energy consumption would be reduced by $4.6\%$ in the AM period and $3.5\%$ in the PM period with corresponding reductions in congestion induced delay of $20\%$ and $14\%$, respectively.   

In future work, we would like to integrate these results with additional alternatives. Eco-routing feedback control selects driving routes based on real-time link estimates of fuel consumption \citep{elbery2018}. Additionally, we would also like to explore techniques for departure time optimization and measure the switching costs associated with carpooling to expand the set of alternatives.

From a social good perspective, \textsc{Copter} is ready for deployment with influenced messages sent through the TripGo  application \citep{tripgo}. The next step is to evaluate the influence model by measuring actual traveler behavior and then feeding those results back into our simulation study. In a deployed scenario, we will be able to integrate additional transportation influence features identified from our survey \citep{mohan2019exploring} (e.g., weather) by updating the acceptability model, if we have enough historical trip behavior, or the influence model by incorporating these features as parameters in the adoption model. Later work will define and assess the impacts of timeliness of a recommendation and compelling messaging by considering Cialdini's influence strategies \citep{cialdini2001harnessing}.

Urban transportation is a huge problem for the environment and citizens' quality of life. \textsc{Copter} demonstrates how multiple AI techniques can be adapted and combined to alleviate congestion and improve the environment through transportation influence.

\section{Acknowledgments}
This work was funded in part by the Advanced Research Projects Agency-Energy (ARPA-E), U.S. Department of Energy, under Award Number DE-AR0000612. The authors would like to thank Filip Dvorak and Aaron Ang for engineering the system, Frances Yan and Victoria Bellotti for contributions to the design and execution of the human studies presented here, and Spenser Anderson for reviewing a draft of this work. The authors appreciate the effort invested by the City of Los Angeles and CalTrans in collecting the data necessary for developing the. Finally, the authors would like to thank the anonymous reviewers for the AAAI/AIES conference on AI, Ethics, and Society and the Journal of AI Research for providing invaluable feedback that has made the writing impactful. 

\printbibliography
\end{document}